# GENERATIVE AI-DRIVEN FORECASTING OF OIL PRODUCTION

Yash Gandhi, Kexin Zheng, Birendra Jha, Ken-ichi Nomura, Aiichiro Nakano, Priya Vashishta, Rajiv K. Kalia

University of Southern California, Los Angeles, CA

## ABSTRACT

Forecasting oil production from oilfields with multiple wells is an important problem in petroleum and geothermal energy extraction, as well as energy storage technologies. The accuracy of oil forecasts is a critical determinant of economic projections, hydrocarbon reserves estimation, construction of fluid processing facilities, and energy price fluctuations. Leveraging generative AI techniques, we model time series forecasting of oil and water productions across four multi-well sites spanning four decades. Our goal is to effectively model uncertainties and make precise forecasts to inform decision-making processes at the field scale. We utilize an autoregressive model known as TimeGrad and a variant of a transformer architecture named Informer, tailored specifically for forecasting long sequence time series data. Predictions from both TimeGrad and Informer closely align with the ground truth data. The overall performance of the Informer stands out, demonstrating greater efficiency compared to TimeGrad in forecasting oil production rates across all sites.

# INTRODUCTION

Forecasting oil and gas production volumes from wells in an oilfield is a classic petroleum engineering problem that has been extensively studied over the years[1,2,3]. Precision in forecasting the produced fluid (oil, gas, and water) volumes plays an important role in determining the accuracy of economic projections, net present value calculation, hydrocarbon reserves estimation, number of wells to be drilled, construction and maintenance of fluid processing facilities, and oil price fluctuations within the petroleum industry[4].

The oil rate forecasting problem can be viewed as a time series forecasting challenge due to the temporal nature of the data, e.g., fluid volume vs. time[6,7]. Several challenges arise in addressing this problem, including an incomplete understanding of the multiphase flow physics that governs the time of water breakthrough in an oil well and the rate of decline in oil production with increasing water production. Additionally, spatial heterogeneity in rock properties, such as permeability and compressibility, further complicates the matter. Nonlinear dependencies between the well flow rate and the reservoir pressure, along with external perturbations induced by wellbore control mechanisms like choke control and hydraulic stimulation or fracturing, often introduce noisy fluctuations in non-stationary time series data[2,5]. Over time, both physics-based reservoir simulation models[8,9] and observation-based empirical and semi-analytical models, such as Arps's power-law model for Decline Curve Analysis (DCA)[10,11], have been developed to forecast oil/gas rates from individual wells or groups of wells within an oilfield.

The two commonly used methods--DCA and reservoir simulation—have their respective limitations. DCA is susceptible to significant extrapolation errors in reservoirs lacking recognizable decline trends, which can occur due to natural or engineered pressure support to the reservoir or changes in well count or quality. On the other hand, the physics-based simulation approach requires information and assumptions about the geologic structure, rock and fluid properties, initial and boundary conditions, and wellbore geometry to forecast production time series[8]. The cost of acquiring such data through well logging, well testing, and seismic surveys can soar into millions of dollars, contingent upon geographical location, well depth, and geologic complexity[9]. Moreover, the usage of these data types in physics-based simulation models often requires years of domain expertise. Due to such bottlenecks in applying the physics-based simulation approach and the need for faster predictive models for real-time optimization and control, there has been intense research into alternative approaches.

Machine learning-based methods have recently emerged as an alternative to reservoir simulation for tackling the production forecasting problem[12,13,14,15,16,17]. Many of these methods rely solely on historical production data (supervised learning) and simpler calculations compared to a simulator, thus yielding significant computational advantages.

Classical time series methods have traditionally served as the backbone for forecasting univariate distributions, such as those seen in oil and gas distributions. These methods typically rely on hand-crafted features and are trained on individual time series. Hyndman et al. offer an extensive review of these methods[18]. Autoregressive integrated moving averages (ARIMA) represent another general class model for forecasting time-series data. ARIMA is well-suited for analyzing stationary stochastic processes with consistent probability distributions. In ARIMA modeling, a stochastic process is characterized by its mean, variance, and autocorrelation function[19,20]. Ayeni and Pilat[21] and Choi et al.[22] applied ARIMA to tackle the challenge of forecasting oil production.

Deep learning models have emerged as superior alternatives to classical methods for end-to-end training in time-series forecasting. Notable examples include recurrent neural networks[17,23,24] and autoregressive moving average methods[25]. Salinas et al. have introduced a deep learning model called DeepAR, specifically designed for forecasting univariate distributions[26]. It leverages an autoregressive recurrent neural network (RNN) to achieve precise probabilistic forecasts following training across diverse time series data. For high-dimensional multivariate forecasting, Salinas et al. have introduced Vec-LSTM, a model that integrates RNN with low-rank covariance matrices. The combination approximates non-Gaussian data, effectively reducing computational complexity[27]. Autoregressive models and normalizing flows have been proposed as effective approaches for learning the time evolution of multivariate probability distributions[28,29,30]. These models have specific requirements, such as constraints on the determinant of the Jacobian in normalizing flows[31].

Here, our focus is on time series forecasting of multivariate distributions for oil and water production from sites encompassing multiple wells across several decades. Inferring the stochastic dependency between past and future values of such lengthy time series containing various trends presents a longstanding challenge in the field. Notable trends include the evolution of multiple wells drilled over time, the underlying decline trends in oil and water rates linked to the physics of multiphase flow in a reservoir, and surface-related controls on wells, such as shut-in events. Numerous challenges in this problem remain unaddressed in previous studies. Many of these studies attempt to model a site's production by combining multiple single-well prediction models or by using coarser time resolutions, e.g., yearly production values instead of monthly production values. However, these approaches often yield poor accuracy because they overlook crucial factors such as the successive addition of new wells and the physics-based interaction among wells at the site. Secondly, models trained solely on predicting either oil or water flow rate, using data for the corresponding phase only, overlook the physics of multiphase flow governing the simultaneous movement of different fluid phases within a reservoir. Thirdly, numerous studies modify/de-noise production data by eliminating large fluctuations, such as zero oil rate values during shut-in events[17], because those methods struggle to learn effectively from noisy datasets. These modifications deprive the model of learning critical physical mechanisms, such as shut-in-induced pressure build-up leading to a surge in the oil rate after a shut-in event. Consequently, inaccurate forecasts occur in real oilfields where shut-in events are prevalent, e.g., 40% of the data could be zero values[17]. Furthermore, excluding such data from the model results in missed opportunities, as shut-in events can offer insights into reservoir permeability and boundaries. Lastly, real-world decision-making under uncertainty necessitates the probability density function (PDF) of predicted oil rates and their evolution over time, a requirement unmet by deterministic modeling methods.

To tackle the challenges in oil rate forecasting, we utilize two generative AI methods: TimeGrad and Informer. TimeGrad[32] is an autoregressive model featuring a tractable likelihood tailored for multivariate distributions. It resembles the seq-to-seq language model proposed by Sutskever et al.[33]. TimeGrad combines a denoising diffusion probabilistic model (DDPM)[34,35,36] with an RNN. On the other hand, Informer[37] is a variation of the transformer architecture. Unlike a standard transformer, it employs sparse probability self-attention to restrict the number of queries.

The novelty of these methods lies in their capability to track the time evolution of the forecast variable's PDF. This enables us to integrate uncertainty and variability in the properties and processes of both subsurface and surface components of an oilfield. We illustrate the applicability and efficacy of TimeGrad and Informer using data from an oilfield that exhibits both high-frequency and low-frequency signals from well operations and reservoir flow physics. These proposed methods learn trends in the data, estimate the evolution of the underlying PDF, and forecast the oil rate with significantly higher accuracy compared to existing methods in the literature.

**RESULTS**

In this section, we describe the forecasts produced by the TimeGrad and Informer models while the results of a standard transformer model are presented in Fig. S3 in the supplementary material.

At each time step, both TimeGrad and Transformer generate 100 samples each. To generate forecasts, we sort the 100 samples in ascending order and select the best value based on quantiles. The median (50$^{th}$ percentile) serves as a robust measure of central tendency, being less affected by extreme values. Lower quantiles (e.g., 25$^{th}$ percentile) offer a conservative estimate for lower bounds of forecasts. Upper quantiles (e.g., 75$^{th}$ percentile) provide insights into higher values and help in evaluating upper bounds. In practical terms, achieving higher accuracy for upper bounds (peaks of oil volume time series) may be crucial compared to lower bounds (troughs of oil volume time series) due to the storage capacity limitations of fluid collection and processing vessels in the oilfield. The generated results correspond to the best quantile forecasts.

We present predictions vs. ground truth oil production for four sites: BEAP, BEAT, EUAT, and EUZE, employing three different generative AI approaches. There are 7,200 timesteps at an interval of 2 days, indicating a frequency of 2 days. In Figs. 1, 2, 3 and 4, panel (A) depicts the observed oil production data at each site. We use an 80:20 split for training and testing across a period of four decades, spanning from 1970 to 2010. Panel (B) illustrates the results of the Informer (IN) model, which predicts 45 timesteps from from March (03) to June (06), 2010 (2010-03 to 2010-06). The IN model utilizes a multivariate distribution across four sites during training. It is trained over 9 epochs with the NLL loss.

Panel (C) showcases the outcomes of the TimeGrad model 1 (TG1), where each site undergoes independent training using the oil and water production data. For additional details regarding water production data, please refer to the supplementary material Figs. S8 and S9. The water production value remains zero for several years at the onset of a site's production life due to the multiphase flow physics governing the reservoir's production behavior[8]. To accommodate this behavior in modeling, we truncate the dataset based on the first occurrence of the non-zero water production (known as the water breakthrough time in the field). Subsequently, we partition the remaining timesteps into an 80:20 ratio for training and testing purposes. Figure S2 in the supplementary material shows the L1 loss for each site.

In panel (D), we present predictions of two distinct TimeGrad models utilizing only oil production values from two pairs of sites: BEAP and BEAT, and EUAT and EUZE. Both pairs exhibit similar oil production patterns. The oil production values for these pairs form the bivariate distribution for the respective models. The forecast spans 45 timesteps from 2010-03 to 2010-06. Models TG1 and TG2 are trained over 40 epochs using a Gated Recurrent Unit (RNN with GRU cell) with L1 loss.

Below, we discuss the forecasts generated by the Informer and the two TimeGrad models across the four sites. Further quantification of these models' performance is provided in Table 1 in the supplementary material. Additionally, we have employed a vanilla transformer to forecast oil production across the four sites, with detailed results presented in Fig. S3 in the supplementary material. Water Production data for each site has also been independently trained using TimeGrad, Informer and Vanilla Transformer and can be seen in the supplementary material: Figs. S4, S5, S6 and S7. Forecasting results for oil production at each site over a period of 6 months are presented in Fig. S11 and Table 2 in the supplementary material.

**BEAP Site:** Figure 1 depicts oil-production forecasts of the Informer (IN) and TimeGrad models TG1and TG2. Panel (A) exhibits 40 years of raw oil production data, with blue indicating training data and orange indicating test data. Panel (B) showcases Informer (IN) results, employing a multivariate distribution for oil production at four sites. The $90^{th}$ percentile is selected based on sample performance, highlighting Informer's strong predictive capability for BEAP's oil rate fluctuations. Panel (C) shows TG1's forecast closely matching the ground truth. The $20^{th}$ percentile is selected, and the deviation across percentiles is minimal. TG1 demonstrates a strong alignment of the forecasted peaks and troughs with the ground truth. Incorporating water production data enhances forecast accuracy, leveraging the interplay between oil and water due to multiphase flow physics. Panel (D) displays TG2's $85^{th}$ percentile forecast, which closely aligns with the ground truth, indicating robust predictive performance for the site. The oil forecast of TG1 is slightly superior to that of TG2 because TG1 can capture higher-order, multiphase flow-based physical mechanisms in the joint oil-water data. These mechanisms play a critical role in contributing to oil production, thus enhancing TG1's forecasting accuracy. It is important to note that TG1's training and forecast periods are shorter or truncated compared to TG2. This is due to the removal of the oil data before the water breakthrough time, a process necessary to accommodate the model's focus on capturing multiphase flow-based physical mechanisms. IN's forecast stands out as the most accurate among the three models owing to its learning capabilities in handling long sequence time series forecasting problems.

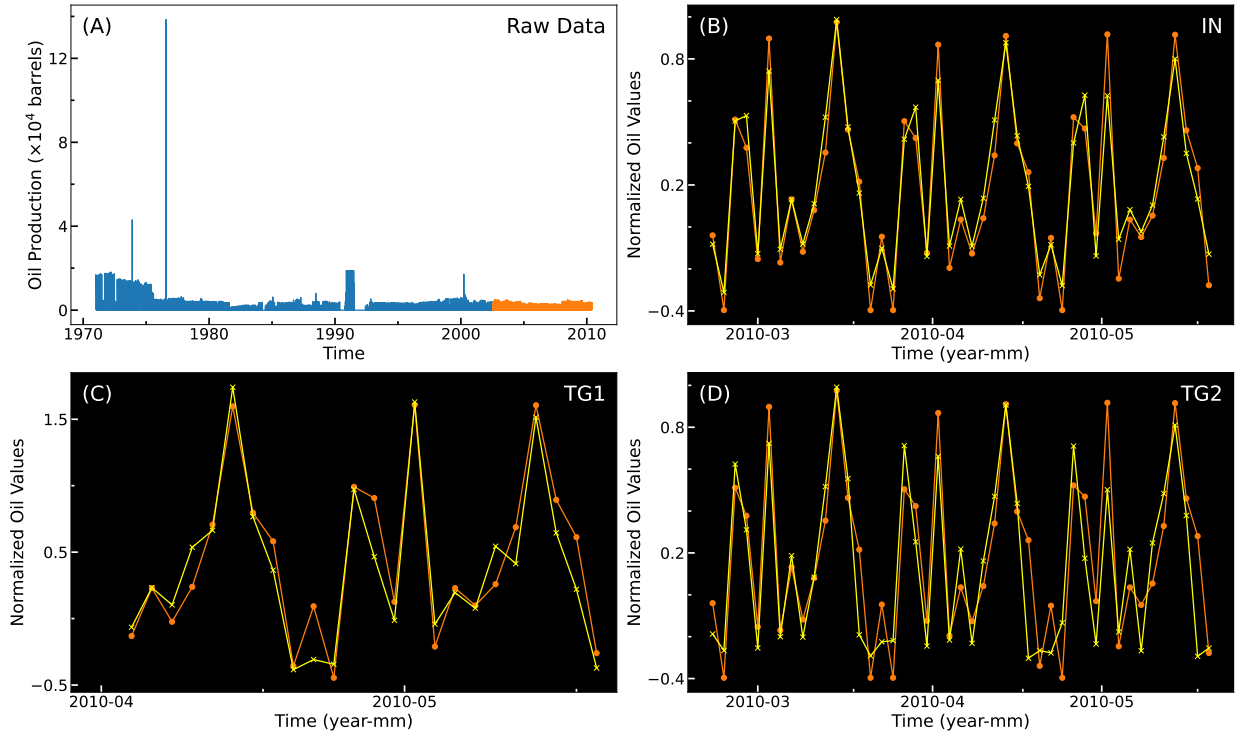

Figure 1: **Oil production forecast for the BEAP site.** In panels (B, C, D), yellow lines are predictions and orange lines are ground truth. Panel (A) presents the raw oil production data, with the blue and orange sections representing the training and test data, respectively. Panel (B) illustrates the results of the Informer (IN) model for BEAP, utilizing a multivariate distribution of oil production across the four sites. Based on the performance of generated samples, we select the 90$^{th}$ percentile. Panel (B) demonstrates the Informer's strong ability to forecast oil rate fluctuations at BEAP. Panel (C) depicts the effective performance of TG1 (20$^{th}$ percentile), with a one-to-one correspondence between peaks and troughs in the ground truth. Utilizing the water production data enhances forecast accuracy by capturing the coupling between oil and water resulting from multiphase flow physics. Panel (D) displays the forecast from the TG2 model (85$^{th}$ percentile), which closely aligns with the ground truth.

**BEAT Site:** The raw data and forecasts for the BEAT site are presented in Fig. 2. In panel (A), the raw oil production values at the BEAT site are displayed, segmented into training (blue) and test (orange) sets. In panel (B), the Informer (IN) forecast at the 50$^{th}$ percentile is showcased, exhibiting peaks and troughs closely aligned with the ground truth. In panel (C), the forecast of the TimeGrad model TG1, which integrates oil and water production, is depicted at the 85$^{th}$ percentile, demonstrating a close alignment of the predictions with the observed data. The TimeGrad model TG2 demonstrates robust performance at the 25th percentile, closely resembling the ground truth, particularly concerning peak values.

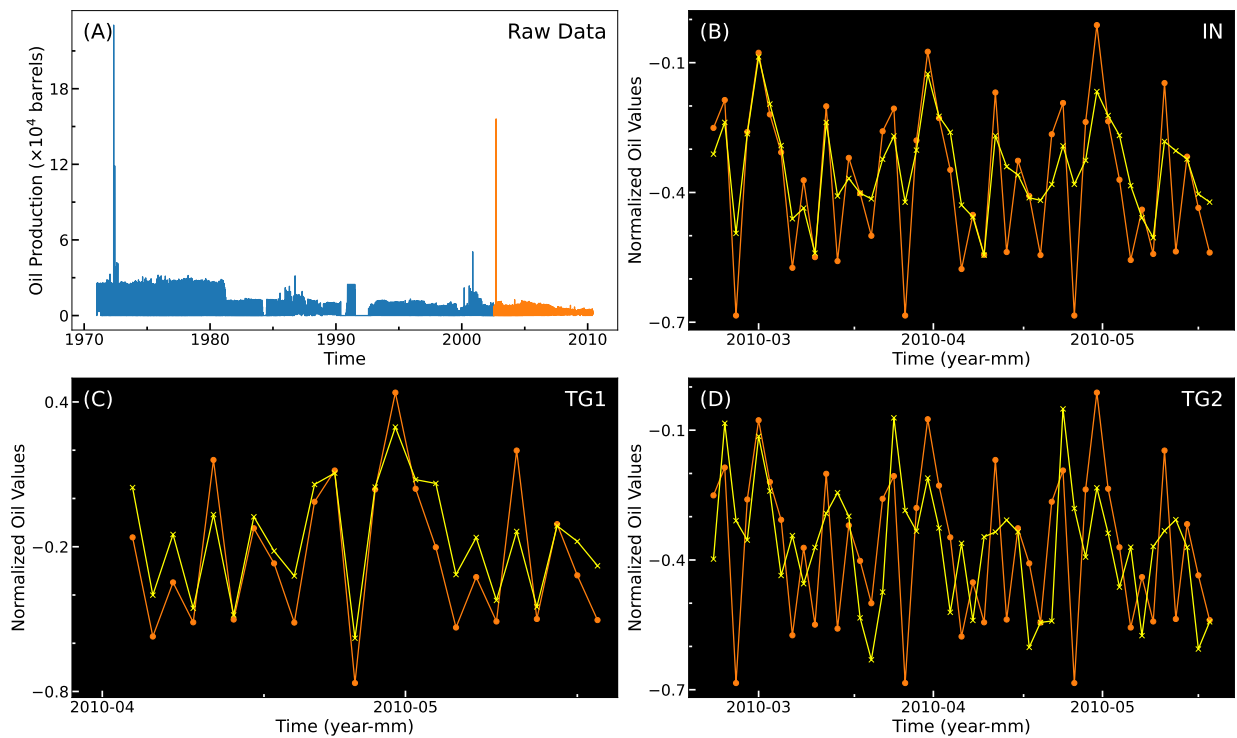

**Figure 2: Oil production forecast for the BEAT site.** In panels (B, C, D), yellow lines represent predictions and orange lines represent ground truth. Panel (A) shows the raw oil production data, divided into training (blue) and test (orange) segments. Panel (B) illustrates the Informer (IN) forecast at the 50th percentile, with the peaks and troughs aligned closely with the ground truth data. Panel (C) displays the forecast of the TimeGrad model TG1 at the 85th percentile, demonstrating a close alignment between the predictions and the observed data. The TimeGrad model TG2 demonstrates strong performance at the 25th percentile.

**EUAT Site:** Figure 3 presents the raw data and forecasts for the EUAT site. Panel (A) displays the raw oil production data, segmented into training (blue) and test (orange) sets. Panel (B). features the Informer (IN) forecast (yellow) at the 70th percentile. Informer accurately captures overall data fluctuations and successfully predicts several peaks in the ground truth data (orange). The non-normalized results for the Informer can be seen in Fig. S10 in the supplementary material.

Yellow lines in panels (C) and (D) represent TG1 forecast at the 20[th] percentile and TG2 forecast at the 75[th] percentile, respectively. Even with the 75[th] percentile, the TG2 forecast fails to align with the forecasts observed at the other three sites. Although the peak positions are reasonably aligned, the peak heights and troughs do not match the ground truth. This discrepancy primarily arises from the significant fluctuations in both the mean and standard deviation of oil production between the training and testing periods in the raw data at EUAT. A comparison between the performance of TG1 and TG2 emphasizes that integrating oil and water production enhances the TimeGrad performance; TG1 captures fluctuations better than TG2, which lacks the water data. Therefore, despite TG1's reduced dataset (where data is truncated at water breakthrough time), TG1 outperforms TG2 due to the additional guidance from the multiphase flow physics governing the oil production process.

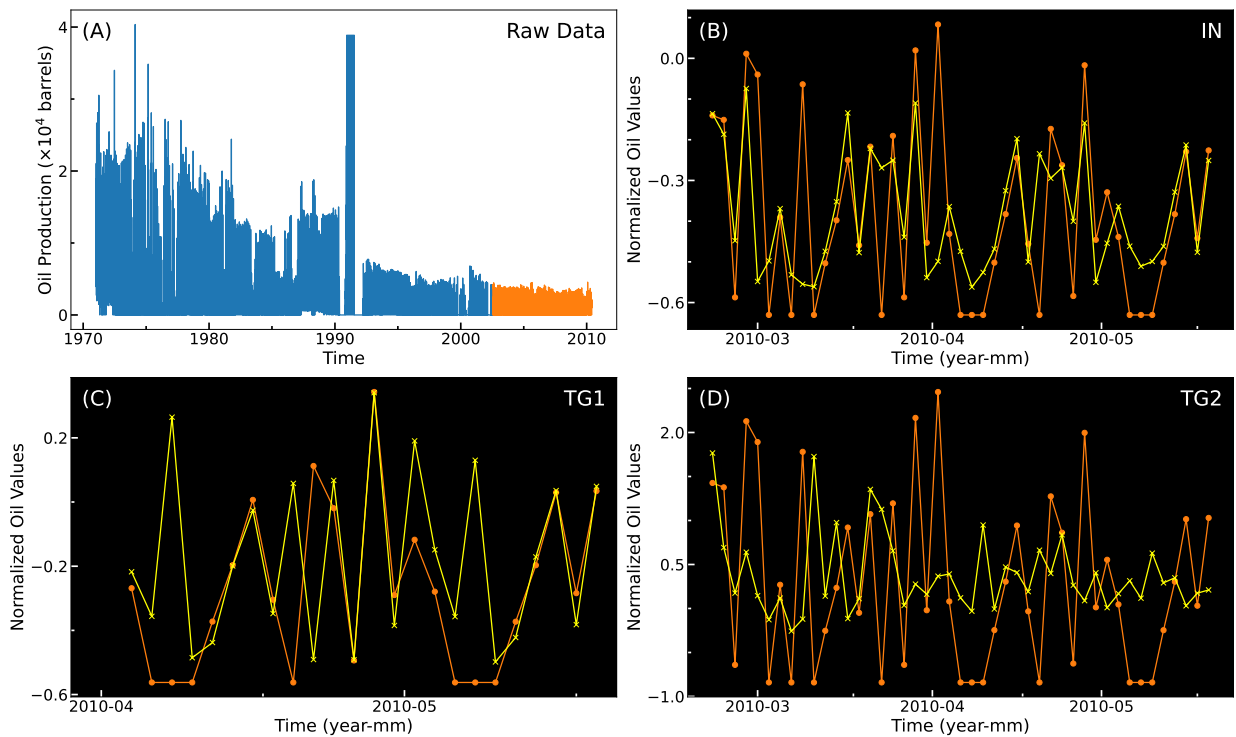

**Figure 3: Oil production forecast for the EUAT site.** In panels (B, C, D), yellow lines represent predictions and orange lines are ground truth. The raw oil production data shown in panel (A) is divided into training (blue) and test (orange) sets. Panel (B) demonstrates the multivariate informer (IN) results at the 70[th] percentile. The Informer

accurately predicts overall data fluctuations and successfully forecasts several peaks. A comparison of models TG1 and TG2 underscores that modeling oil and water production together significantly enhances the performance of the TimeGrad model. The TG2 forecast, even at the 75$^{th}$ percentile, does not align with the forecasts observed at the other three sites. While the peak positions in time demonstrate reasonable agreement, the peak and trough amplitudes do not match the ground truth. This disparity primarily arises from significant fluctuations in the mean and standard deviation of EUAT oil production between the training and testing periods. Despite having a smaller dataset, the TimeGrad model TG1 effectively predicts fluctuations, especially at the 20$^{th}$ percentile shown here.

**EUZE Site:** In Fig. 4, panel (A) displays the raw oil production data spanning from 1970 to 2010, with the blue segment indicating training data and the orange segment indicating test data. Panel (B) displays the forecast of the Informer (yellow) and ground truth (orange) from March to June 2010. At the 90$^{th}$ percentile, the predicted peaks and troughs align almost perfectly with the ground truth. The non-normalized results for the Informer are shown in the supplementary material Fig. S10. Panel (C) indicates the performance of the TG1 model at the 30$^{th}$ percentile. The predicted peaks and troughs align well with the ground truth, except for the merging of peaks in mid-April and May. This is caused by sparse water production during this decade at EUZE; please see supplementary material Fig. S7 for water data. Panel (D) illustrates that the TG2 model forecast can effectively capture the uneven oil production pattern, particularly at the 65$^{th}$ percentile.

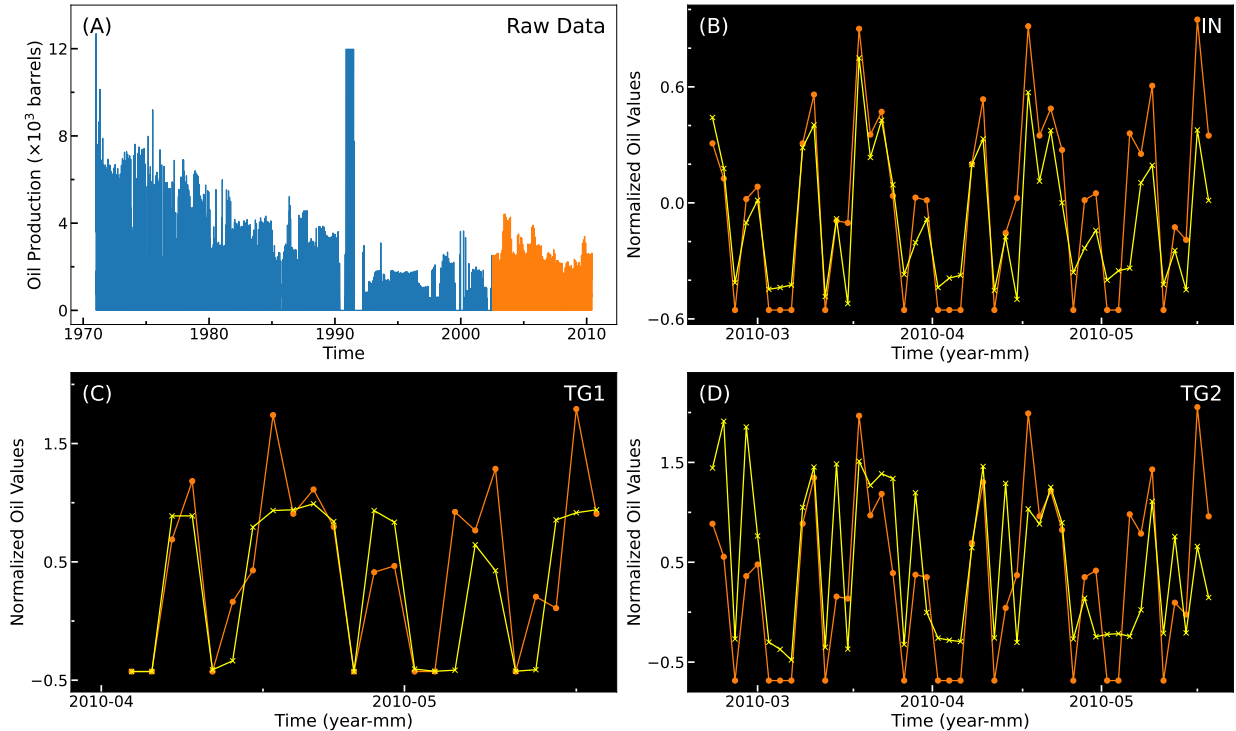

**Figure 4: Oil production forecast for the EUZE site.** In panels (B, C, D), yellow lines represent predictions and orange lines are ground truth. Panel (A) shows the raw oil production data from 1970 to 2010. The blue and orange segments represent training and test data, respectively. Panel (B) illustrates the performance of the Informer (IN). At the 90$^{th}$ percentile, the predicted peaks and troughs (yellow) are almost perfectly aligned with the ground truth (orange). The TG1 model performs well at the 30$^{th}$ percentile (panel (C)). The predicted peaks and troughs generally align with the ground truth, except for mid-April and May, where the peaks appear merged. Sparse water production data during the initial water production period (see supplementary material Fig. S7) compromises the training quality. Panel (D) demonstrates the relatively strong performance of the TG2 model in capturing the uneven oil production pattern, particularly at the 65$^{th}$ percentile.

## CONCLUSION

Forecasting fluid production at oilfield sites with multiple wells and multiphase flow (simultaneous flow of oil, water, and gas phases) is a problem of huge economic and scientific significance, applicable to petroleum recovery, groundwater and geothermal energy extraction, and energy storage technologies. The complexity inherent in the production data, the governing physics, and the uncertainty surrounding physical properties and human controls render the problem particularly challenging. In response, we suggest employing generative AI methods, specifically leveraging TimeGrad and Informer techniques, for forecasting oil production. We demonstrated the predictive power of the methods across four sites, each exhibiting a diverse range of data variability. In this comparison, Informer consistently outperforms TimeGrad in terms of accuracy and efficiency when forecasting oil production rates across all sites. Our analysis reveals that the multivariate framework inherent in our methods enables us to capture the physical coupling between oil and water production processes, leading to enhanced accuracy. Unlike deterministic approaches such as PDE-based simulations, our probabilistic framework effectively models the temporal evolution of uncertainty and PDF of the predicted oil production. This allows our models to forecast at desired percentiles, enabling tailored accuracy levels across different segments of the time series. For instance, in field operations, prioritizing higher accuracy in peak oil periods (reflecting the timing and magnitude of oil production surges) over troughs is often crucial due to its impact on storage capacity and revenue generation.

The AI-based forecasting framework presented here holds promise for various applications beyond oil production forecasting. One such critical area is the study of earthquakes induced by the extraction and injection of fluids in underground reservoirs, a problem of immense societal and scientific importance[38]. Induced earthquakes have been linked to various activities such as wastewater disposal, geologic carbon sequestration, hydraulic fracturing, extraction of geothermal energy, hydrocarbons, and groundwater[39,40,41]. Earthquake risk assessment and management initiatives have prompted the implementation of real-time monitoring for ground surface deformation near numerous oil, gas, geothermal, and groundwater sites worldwide. The time series data of ground deformation displays complex trends associated with the well activity and the multiphysics coupling between fluid flow and faulting/fracturing processes. This complexity makes PDE-based simulation of induced earthquakes computationally intractable. Generative AI-based probabilistic forecasting frameworks, like TimeGrad and Informer as presented here, show promise because they can effectively incorporate the uncertainty and variability in the properties and processes of surface and subsurface components within an induced earthquake system.

**METHODS**

The overarching problem we aim to model is as follows: Given a specific input, $(x_1^t, \ldots, x_{L_x}^t \mid x_i^t \in R^{d_x})$ at time $t$, predict the corresponding sequence $(y_1^t, \ldots, y_{L_x}^t \mid y_i^t \in R^{d_y})$. We employ three different methodologies to forecast oil rate production from sites with numerous wells:

(1) TimeGrad: This approach utilizes a denoising diffusion probabilistic model (DDPM) coupled to a recurrent neural network (RNN) to model multivariate time series data.

(2) Vanilla transformer: A basic transformer model is employed for forecasting oil rate production rates over time at various sites.

(3) Informer: This method employs an efficient transformer architecture tailored for forecasting long sequence time series forecasting (LSTF).

These methods are briefly described below, with additional explanations available in the supplementary material.

## A. TIMEGRAD

TimeGrad is an autoregressive method, designed to predict the temporal patterns of a multivariate probability distribution. It has demonstrated successful applications in forecasting various datasets, including traffic, stock exchange, electricity, solar, and Wikipedia open datasets. TimeGrad comprises a DDPM coupled to an RNN, as illustrated in the left panel of Fig. 5. The detailed algorithm is given in the supplementary material, Fig. S1.

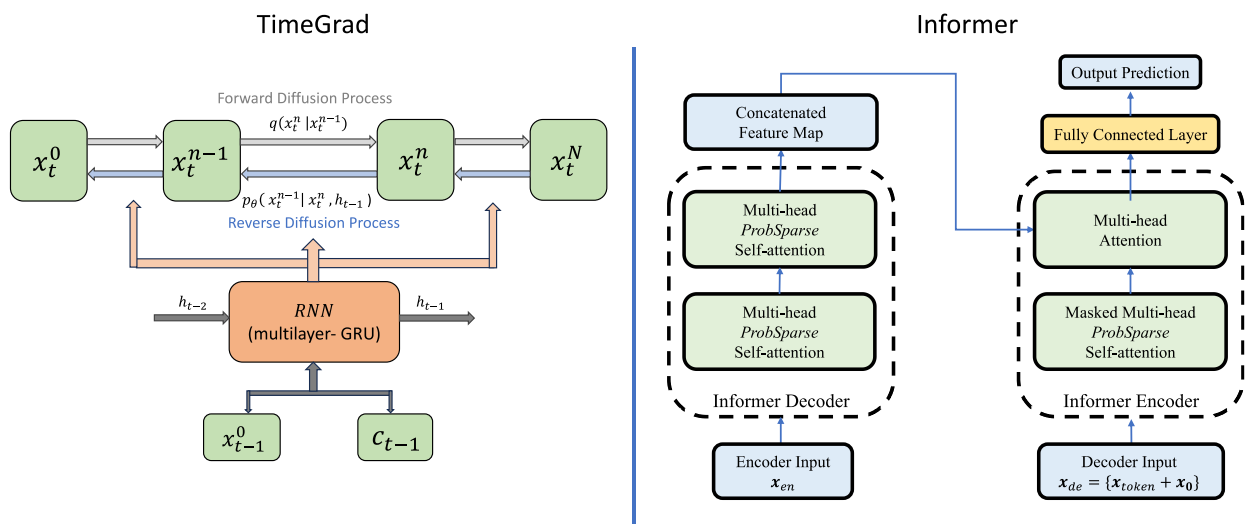

**Figure 5:** The left panel illustrates the TimeGrad architecture[32], which comprises a DDPM connected to an RNN. During the forward process, Gaussian noise is introduced to the conditional distribution $q(x_n|x_{n-1})$ while the reverse process involves the denoising of the parameterized distribution $p_\theta(x_{n-1}^t|x_n^t, h_{t-1})$ at time *t*. This denoising process incorporates input from the hidden state of the RNN at time step *t*-1. The right panel shows the encoder and decoder architectures of the Informer[37]. Both components feature *ProbSparse* in their multi-head attention mechanisms, enhancing the efficiency of the Informer compared to a vanilla transformer.

### A-1. Denoising Diffusion Probabilistic Model (DDPM)

The DDPM draws inspiration from the nonequilibrium physics of diffusive processes[34]. It serves as a generative model for a probability distribution $q(x_0)$ of input data $x_0$. The DDPM has demonstrated effectiveness in modeling various data types, including images, audio, and text-to-speech synthesis. The DDPM produces higher quality images and text compared to other generative models like the variational autoencoder (VAE), and it exhibits greater stability than the generative adversarial network (GAN).

The DDPM is trained to approximate $q(x_0)$ using a parameterized distribution, $p_\theta(x_0)$, through latent space variables $(x_0, x_1, \ldots, x_N)$ of the same dimension as $x_0$:

$$p_\theta(x_0) = \int p_\theta(x_0, x_1, \ldots, x_N) \, dx_1, \ldots, dx_N \tag{1}$$

The DDPM encompasses a forward diffusive process and a reverse denoising process, as illustrated schematically in the left panel of Fig. 5(a). In the forward process, the posterior $q(x_1, \ldots, x_N | x_0)$ is represented as a Markov chain over the latent variables,

$$q(x_1, \ldots, x_N | x_0) = \prod_{n=1}^{N} q(x_n | x_{n-1}), \tag{2}$$

and noise is gradually introduced to the conditional distribution during the forward process, resulting in incremental increase in variance at each step. After a significant duration, the probability distribution, $q(x_N)$, at the horizon transitions into a tractable Gaussian distribution.

The reverse denoising process operates a Markov chain for the approximator $p_\theta(x_0, \ldots, x_N)$:

$$p_\theta(x_0, \ldots, x_N) = p(x_N) \prod_{n=N}^{1} p_\theta(x_{n-1} | x_n), \tag{3}$$

where $p(x_N) = N(x_N; 0, I)$ is the Gaussian noise. The reverse process entails iterative denoising of the distribution, ultimately yielding the approximator, $p_\theta(x_0)$, for the data distribution $q(x_0)$. The parameters $\theta$ are shared and learned by minimizing the negative of the log likelihood to align with the data distribution $q(x_0)$: $\min_\theta E_{q(x_0)}[-\log p_\theta(x_0)]$. Further details are provided in appendix 1 of the supplementary material.

### A-2. Coupling between DDPM and RNN

For a multivariate time series $x_{0,t}$, where $t$ is the time index and $x_0$ comprises $D$ elements (1, 2,..., $D$) at each time step, TimeGrad divides the entire time sequence $t = [1, T]$ into a contiguous context interval $[1, t_0 - 1]$ and a contiguous prediction interval $[t_0, T]$. The multivariate probability distribution for the prediction level learns from the conditional distribution over the context and the past prediction levels:

$$q(x_{t_0,T}|x_{1,t_0-1}) = \prod_{t=t_0}^{T} q(x_t|x_{1:t-1}), \tag{4}$$

Each conditional distribution on the right-hand side of Eq. (4) is calculated using the DDPM model and the temporal dynamics up to time $t$ through an RNN; further details are provided in the supplementary material.

### A-3. Training and Forecasting

Our oil, gas, and water production data are sourced from four sites operational between 1971 and 2012. We select fifteen wells from each site. From the time series data, we randomly sample from a context and an adjacent prediction window and optimize the negative log likelihood,

$$-\sum_{t=t_0}^{T} \log p_\theta(x_{0,t}|h_{t-1}), \tag{5}$$

with respect to the parameters $\theta$ in the context window using an RNN. Further details are available in the supplementary material.

### B. TRANSFORMERS

We also employ a vanilla transformer and Informer to predict oil production from each site. Transformers are potent generative AI tools used for modeling language, vision, and audio data. They have been successfully adapted to various sequence-to-sequence modeling tasks, such as the decision transformer for offline reinforcement learning[42], and energy transformers[43] for tasks like data imputation in images, graph classification and graph anomaly detection.

A transformer comprises an encoder and a decoder, both employing four fundamental operations: attention, feed-forward (FF) multilayer network, residual connection, and layer normalization. Numerous variations of transformers exist based on how these four basic operations are combined. For instance, attention and FF layers can be interleaved, or attention operations may precede the addition of FF layers towards the end, or FF layers can be inserted before the attention layers within each transformer block. More advanced transformer architectures have been designed using neural network-based search methods[44]. Despite extensive research, there remain open questions regarding the roles of the four basic operations: what constitutes the optimal combination of these operations within a transformer block, and how essential is the attention mechanism in learning long-range dependencies?

The central quantity in a transformer is self-attention, which involves a tuple: query ($Q \in R^{L_Q \times d}$), key ($K \in R^{L_k \times d}$), and value ($V \in R^{L_v \times d}$). In a vanilla transformer, self-attention is computed from,

$$A(Q, K, V) = \text{Softmax}\left(\frac{QK^T}{\sqrt{d}}\right) V, \qquad (6)$$

where $d$ is the input dimension. If $q_i$, $k_i$, and $v_i$ are the $i^{th}$ elements of $Q$, $K$ and $V$, the attention of the $i^{th}$ query can be written as,

$$A(q_i, K, Q) = \sum_j \frac{\kappa(q_i, k_j)}{\sum_l \kappa(q_i, k_l)} v_j = E_{p(k_j|q_i)}[v_j], \qquad (7)$$

where $(k_j|q_i) = \frac{\kappa(q_i,k_j)}{\sum_l \kappa(q_i,k_l)}$ and $\kappa(q_i, k_j) = \exp(q_i k_j / \sqrt{d})$. The vanilla transformer is known to be inefficient for LSTF because computing attention for the $i^{th}$ query demands quadratic computation, and memory usage scales as $O(L_Q L_K)$.

Numerous sparse transformers have been introduced to overcome the limitations of a vanilla transformer in forecasting long sequence time series. These include heuristic methods aimed at mitigating the quadratic dot-product computational complexity[45,46]; $O(L \log L)$ Reformer[47] for very long sequences; linear scaling Linformer[48]; and the Compressive Transformer[49] capable of handling LSTF, yet it struggles to alleviate the quadratic dot-product complexity inherent in the attention mechanism; and Transformer-XL[50].

We use Informer, specifically designed for LSTF. It employs sparse probability self-attention to constrain the number of queries to the top $u$ queries. Consequently, the self-attention simplifies to,

$$A(\boldsymbol{Q}, \boldsymbol{K}, \boldsymbol{V}) = \mathrm{Softmax}\left(\frac{\overline{\boldsymbol{Q}}\boldsymbol{K}^T}{\sqrt{d}}\right)\boldsymbol{V}, \qquad (8)$$

where $\overline{\boldsymbol{Q}}$ is a sparse matrix whose elements are the top $u$ queries picked from a sparsity function,

$$M(q_i, K) = \log \sum_{j=1}^{L_K} \exp\left(\frac{q_i k_j}{\sqrt{d}}\right) - \frac{1}{L_K} \sum_{j=1}^{L_K} \exp\left(\frac{q_i k_j}{\sqrt{d}}\right). \qquad (9)$$

The top $u$ queries are determined by setting $u = c \log L_Q$, where $c$ is a sampling constant. This decreases the dot-product computation to $O(\log L_Q)$ for each query and reduces the layer memory usage to $O(L_K \log L_Q)$. Another notable feature of the Informer is its ability to output long sequences with just one forward step, thus avoiding cumulative errors during inference. The right panel in Fig. 5 schematically illustrates the Informer architecture. Detailed description of the encoder and decoder architectures can be found in the supplementary material. The loss function employed by the Informer is negative log likelihood (NLL) with respect to the target sequences. The NLL at the output of the decoder is then propagated back through the entire model.

**ACKNOWLEDGMENTS**

R.K.K. acknowledges support from the Center for Integrated Nanotechnologies, an Office of Science User Facility operated for the U.S. Department of Energy, Office of Science. Sandia National Laboratories is a multi-mission laboratory managed and operated by National Technology & Engineering Solutions of Sandia, LLC, a wholly owned subsidiary of Honeywell International, Inc., for the U.S. DOE's National Nuclear Security Administration under contract DE-NA-0003525. R.K.K. would also like to acknowledge the support of the Ershaghi Center for Energy Transition (E-CET) at the University of Southern California.

# SUPPLEMENTARY INFORMATION

## APPENDIX 1: METHODS

### A. TIMEGRAD

Here we describe the integration of DDPM[1] and RNN within TimeGrad.

### A-1. Denoising Diffusion Probabilistic Model (DDPM)

In the forward process, the conditional distribution involves a gradual increment of Gaussian noise,

$$q(x_n|x_{n-1}) = N(x_n; \sqrt{1-\beta_n}\, x_{n-1}, \beta_n I), \text{ and } \beta_{n+1} > \beta_n. \tag{S1}$$

and in the reverse denoising step, the parameterized distribution is defined by:

$$p_\theta(x_{n-1}|x_n) = N(x_{n-1}; \mu_\theta(x_n, n), \Sigma_\theta(x_n, n)). \tag{S2}$$

The parameters $\theta$ are shared and learned by minimizing the negative of the log likelihood to fit the data distribution,

$$min_\theta\, E_{q(x_0)}\left[-\log p_\theta(x_0)\right]. \tag{S3}$$

Using Eq. (S2) and Jensen's inequality, Eq. (S3) can be expressed as

$$E_{q(x_0)}\left[-\log p_\theta(x_0)\right] \leq E_{q(x_{0:N})}\left[-\log p_\theta(x_{0:N}) + \log q(x_{1:N}|x_0)\right], \tag{S4}$$

The goal is to minimize the upper bound of Eq. (S4):

$$min_\theta\, E_{q(x_0)}\left[-\log p(x_N) + \log \frac{q(x_{1:N}|x_0)}{p_\theta(x_{0:N})}\right], \tag{S5}$$

which can be recast as

$$min_\theta\, \left[-\log p_\theta(x_0|x_1) + D_{KL}(q(x_N|x_0)||p(x_N)) + \sum_{n=1}^{N} D_{KL}(q(x_{n-1}|x_n, x_0)||p_\theta(x_{n-1}|x_n))\right]. \tag{S6}$$

where $D_{KL}$ is the Kullback-Liebler divergence and $q(x_{n-1}|x_n, x_0)$ is a Gaussian distribution:

$$q(x_{n-1}|x_n, x_0) = N(x_{n-1}|\tilde{\mu}_n(x_n, x_0), \tilde{\beta}_n I), \tag{S7}$$

with the mean and divergence given by,

$$\tilde{\mu}_n(x_n, x_0) = \frac{\sqrt{\alpha_n}(1-\sqrt{\tilde{\alpha}_{n-1}})}{(1-\tilde{\alpha}_n)} x_n + \frac{\sqrt{\tilde{\alpha}_n}\beta_n}{(1-\tilde{\alpha}_n)} x_0; \tag{S8}$$

$$\tilde{\beta}_n = \frac{(1-\tilde{\alpha}_{n-1})}{(1-\tilde{\alpha}_n)} \beta_n. \tag{S9}$$

The KL divergence in Eq. (S6) becomes,

$$D_{KL}(q(x_{n-1}|x_n, x_0)||p_\theta(x_{n-1}|x_n)) = E_q\left[\frac{1}{2\Sigma_\theta}||\tilde{\mu}_n(x_n, x_0) - \tilde{\mu}_\theta(x_n, n)||^2\right] + C, \tag{S10}$$

where
$$\tilde{\mu}_\theta(x_n, n) = \frac{1}{\sqrt{\tilde{\alpha}_n}}\left(x_n - \frac{\beta_n}{\sqrt{1-\tilde{\alpha}_n}}\epsilon_\theta(x_n, n)\right), \quad \epsilon \sim N(0, I) \tag{S11}$$

$$x_n(x_0, \epsilon) = \sqrt{\tilde{\alpha}_n}\, x_0 + (1 - \tilde{\alpha}_n)\epsilon, \tag{S12}$$

and $C$ is a constant. Thus, the minimization objective simplifies to,

$$E_{x_0,\epsilon}\left[\frac{\beta_n^2}{2\Sigma_\theta \alpha_n(1-\tilde{\alpha}_n)}\,||(\epsilon - \epsilon_\theta(\sqrt{\tilde{\alpha}_n}\, x_0 + \sqrt{1-\tilde{\alpha}_n}\,\epsilon, n))||^2\right], \tag{S13}$$

which is similar to the loss in score matching[2]. After training, Langevin dynamics can be used for sampling:

$$x_{n-1} = \frac{1}{\sqrt{\alpha_n}}\left(x_n - \frac{\beta_n}{\sqrt{1-\tilde{\alpha}_n}}\epsilon_\theta(x_n, n)\right) + \sqrt{\Sigma_\theta}\, z, \tag{S14}$$

where $z \sim N(0, I)$ for $n = [2, N]$ and $z = 0$ for $n = 1$.

### A-2. Coupling between DDPM and RNN

In the manuscript, we elaborate on the computation of conditional distributions as described in Eq. (S4), and we elucidate how an RNN architecture captures the temporal dynamics up to time $t$. The hidden state $h_t$ of the RNN is updated by,

$$h_t = RNN_\theta(\text{concat}(x_{0,t}), h_{t-1}). \tag{S15}$$

with the initial hidden state $h_0 = 0$. The $RNN_\theta$ is a multilayer LSTM or GRU with shared weights $\theta$. The right-hand side of Eq. (4) in the manuscript is approximated by,

$$\prod_{t=t_0}^{T} p_\theta(x_{0,t}|h_{t-1}), \tag{S16}$$

where $\theta$ denotes the combined weights of the DDPM and RNN.

## A-3. Training and Forecasting

Following the derivation of Eq. (S13), it can be demonstrated that the optimization described in Eq. (5) of the manuscript simplifies to

$$E_{x_0,\epsilon}\left[||(\epsilon - \epsilon_\theta(\sqrt{\tilde{\alpha}_n}\, x_0 + \sqrt{1-\tilde{\alpha}_n}\, \epsilon, \mathbf{h}_{t-1}, n))||^2\right], \tag{S17}$$

when the variance is chosen to be $\Sigma_\theta = \tilde{\beta}_n$.

To forecast oil in the prediction window, we run RNN over the dataset to obtain $\mathbf{h}_T$ from Eq. (S15) and subsequently sample from the following algorithm,

$$\mathbf{x}_{n-1,t} = \frac{1}{\sqrt{\alpha_n}}\left(\mathbf{x}_{n,t} - \frac{\beta_n}{\sqrt{1-\tilde{\alpha}_n}}\epsilon_\theta(\mathbf{x}_{n,t}, \mathbf{h}_{t-1}, n)\right) + \sqrt{\Sigma_\theta}\, \mathbf{z}, \tag{S18}$$

to obtain $\mathbf{x}_{0,T+1}$ at time step $T+1$ and hidden state $\mathbf{h}_{T+1}$ by running the RNN. This procedure is repeated until we reach the end of the prediction window. Starting from $\mathbf{h}_T$, we repeatedly sample to obtain quantiles representing the uncertainty in our prediction of oil and water rates.

Lastly, it is worth noting that we normalize the data within each context window to handle varying time scales. This involves subtracting the mean of the context window from each time series and dividing by the variance of the context window before inputting the time series into the model. Additionally, we restore the original scale before making inferences. Figure S1 illustrates the TimeGrad algorithm.

---

**Algorithm 1:** Training for each time series step $t \in [t_0, T]$

**Input:** data $\mathbf{x}_t^0 \sim q_x(\mathbf{x}_t^0)$ and state $\mathbf{h}_{t-1}$
**repeat**
  Initialize $n \sim \text{Uniform}(1, \ldots, N)$ and $\epsilon \sim N(\mathbf{0}, \mathbf{I})$
  Take gradient step on

  $$\nabla_\theta\, ||\epsilon - \epsilon_\theta(\sqrt{\tilde{\alpha}_n}\mathbf{x}_t^0 + \sqrt{1-\tilde{\alpha}_n}\, \epsilon, \mathbf{h}_{t-1}, n)||^2$$

**until** converged

---

**Algorithm 2:** Sampling $\mathbf{x}_t^0$ via annealed Langevin dynamics

**Input:** noise $\mathbf{x}_t^N \sim N(\mathbf{0}, \mathbf{I})$ and state $\mathbf{h}_{t-1}$
**for** $n = N$ to $1$ **do**
  **if** $n > 1$ **then**
    $\mathbf{z} \sim N(\mathbf{0}, \mathbf{I})$
  **else**
    $\mathbf{z} = \mathbf{0}$
  **end if**

  $$\mathbf{x}_t^{n-1} = \frac{1}{\sqrt{\alpha_n}}\left(\mathbf{x}_t^n - \frac{\beta_n}{\sqrt{1-\tilde{\alpha}_n}}\epsilon_\theta(\mathbf{x}_t^n, \mathbf{h}_{t-1}, n)\right) + \sqrt{\Sigma_\theta}\, \mathbf{z}$$

**end for**
**Return:** $\mathbf{x}_t^0$

**Figure S1:** In TimeGrad, we use Algorithm 1[3] at each time step for training within the prediction window and Algorithm 2[3] for sampling via annealed Langevin dynamics.

### B.  TRANSFORMER

The Transformer model undergoes training for 40 epochs using NLL loss. It encompasses 3 encoder and decoder layers with a dropout rate of 0.2. Model training utilizes AdamW optimizer with an initial learning rate of $10^{-4}$. We also adjust the number of layers, the learning rate, and the dimensions of the model to assess their impact on the results. We find these changes have negligible effects on oil/water forecasts.

### C.  INFORMER

The Informer employs an encoder-decoder architecture. In the encoder, a distilling operation is used to prioritize the dominant features, enabling the creation of a focused self-attention feature map. At the $t^{\text{th}}$ sequence input in layer $j$, denoted as $X_j^t$, the attention block is traversed, incorporating Multi-head ProbSparse self-attention. The ProbSparse self-attention is defined as follows:

$$A(Q, K, V) = \text{Softmax}\left(\frac{\overline{Q}K^T}{\sqrt{d}}\right) V, \quad (S19)$$

where $Q$, $K$, $V$ and $d$ represent query, key, value, and input dimension, respectively.

The $t^{\text{th}}$ sequence input in the next layer $j+1$, denoted as $X_{j+1}^t$, is obtained by applying a 1-D convolutional filter ($\text{Conv1d}(\cdot)$), followed by an activation function ($\text{ELU}(\cdot)$), and a max-pooling layer. The distilling operation can be expressed as:

$$X_{j+1}^t = \text{MaxPool}(\text{ELU}(\text{Conv1d}([X_j^t]_{\text{AB}}))). \quad (S20)$$

The Encoder contains multiple copies of the main stack, with inputs halved to ensure the robustness of the distilling operation. The output of all stacks is concatenated and collectively considered as the final hidden representation of the encoder.

The decoder is a standard architecture, constructed by a stack of two identical multi-head attention layers. In the decoder, a masked multi-head attention is applied by setting masked dot-products as $-\infty$ when computing ProbSparse self-attention. To address inefficiencies in long prediction, generative inference is applied. This involves creating the input of the decoder as a concatenated vector:

$$X_{de}^t = \text{Concat}(X_{token}^t, X_0^t), \qquad (S21)$$

where $X_{token}^t \in \mathbb{R}^{L_{token} \times d_{model}}$ is the start token and $X_0^t \in \mathbb{R}^{L_y \times d_{model}}$ is a placeholder for the target sequence, which contains target sequ1ence's timestamp.

**APPENDIX 2: RESULTS**

**TimeGrad Model TG1**

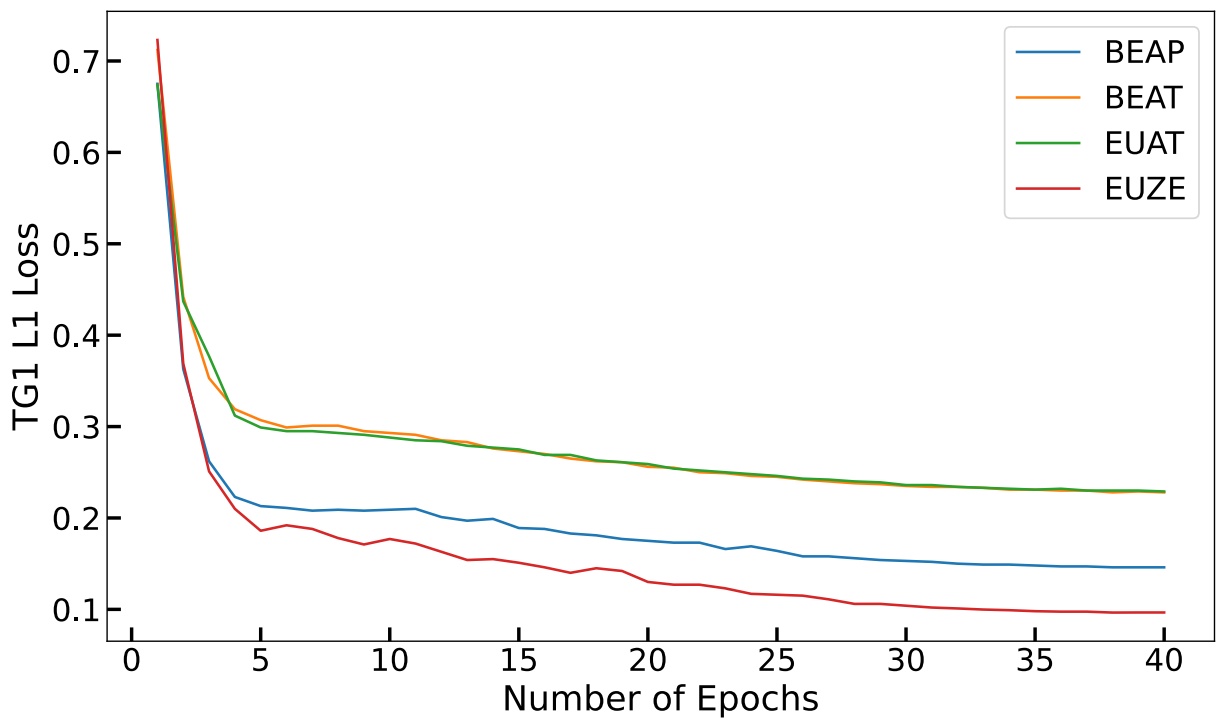

**Figure S2:** L1 loss over 40 epochs for BEAP, BEAT, EUAT and EUZE sites. In TG1, we train the TimeGrad model independently for each site using oil and water production. The loss steadily

decreases within the first 20 epochs. However, beyond 35 epochs, there is minimal change observed in the loss.

**Vanilla Transformer**

Here are the results of the transformer model. While the forecasts are impressive, the loss function indicates overfitting. Unlike the vanilla transformer, the Informer does not overfit.

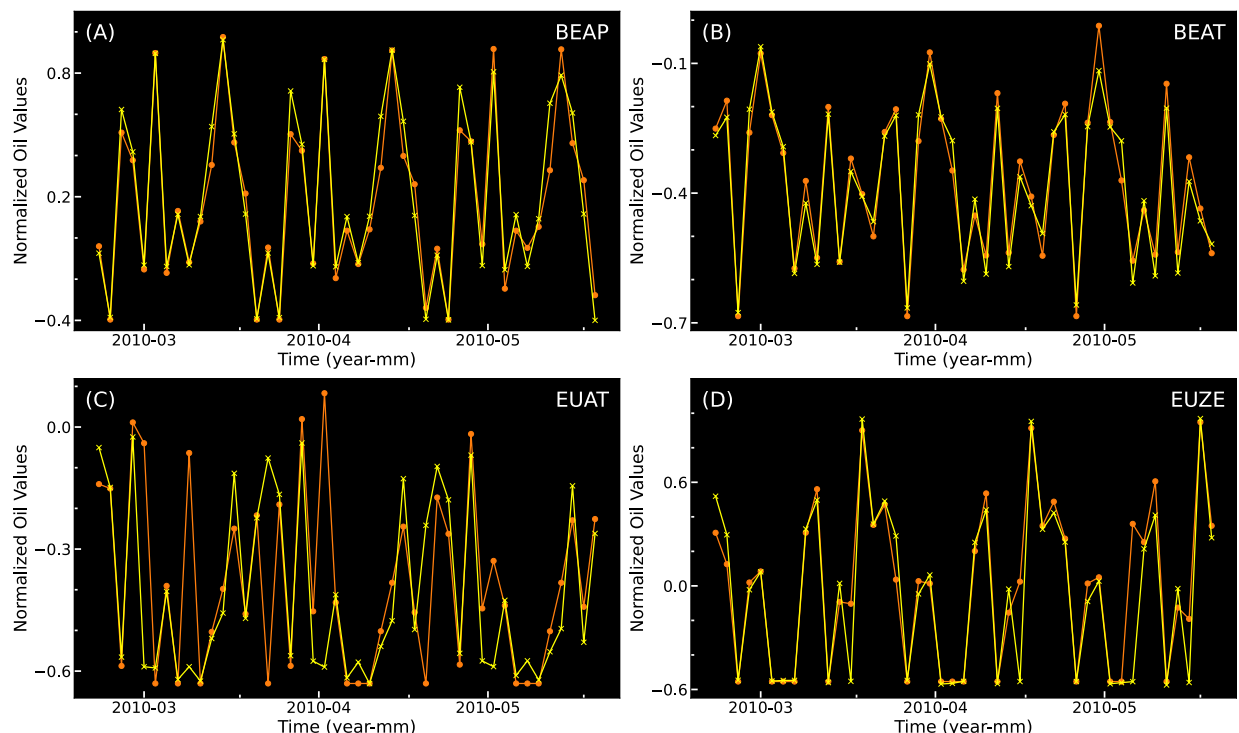

**Figure S3:** Oil production forecasts for the four sites are generated independently. The vanilla transformer forecasts 45 timesteps from March to June 2010, with the 50$^{th}$ percentile chosen for all sites. The transformer forecasts outperform those of TimeGrad and Informer. Notably, the forecasts closely match the ground truths for BEAP, BEAT, and EUZE (panel (A, B, D)). In panel (C), the vanilla transformer demonstrates superior performance in predicting peaks and troughs of the timeseries compared to the Informer model.

**APPENDIX 3: WATER PRODUCTION FORECAST**

Here, we present the predictions of TimeGrad and Informer compared to the ground truth for water production at the four sites. In Figures S4, S5, S6 and S7, panel (A) displays the raw water production volume data. Due to numerous zero values in the water data during the initial stages of oil production—a result of multiphase flow physics in porous media (Aziz and Settari, 1979)—we eliminate all consecutive zeros from the starting date. The modeling of water production commences with the first non-zero value at each site. The water production data spans 1990-10-22 to 2010-06-03 at BEAP, from 1987-09-08 to 2010-06-03 at BEAT, 1986-03-25 to 2010-06-03 at EUAT, and 1987-12-01 to 2010-06-03 at EUZE. We use an 80:20 data split for training and testing. For each site, we combine oil and water production data for forecasting from April 2010 to June 2010.

Panel (B) depicts the results of the Informer (IN) model, where BEAP, BEAT and EUZE are trained for 20 epochs, while EUAT for 10 epochs. Panel (C) displays the results of the TimeGrad model 1 (TG1). Panel (D) exhibits the predictions of the Vanilla Transformer (TR). All sites undergo training for 20 epochs with NLL loss. Notably, BEAP exhibits minimal water production data during the training period compared to the testing period, resulting in inconsistent prediction. However, other sites demonstrate relatively satisfactory performances, resulting from more substantial water production data during the training phase.

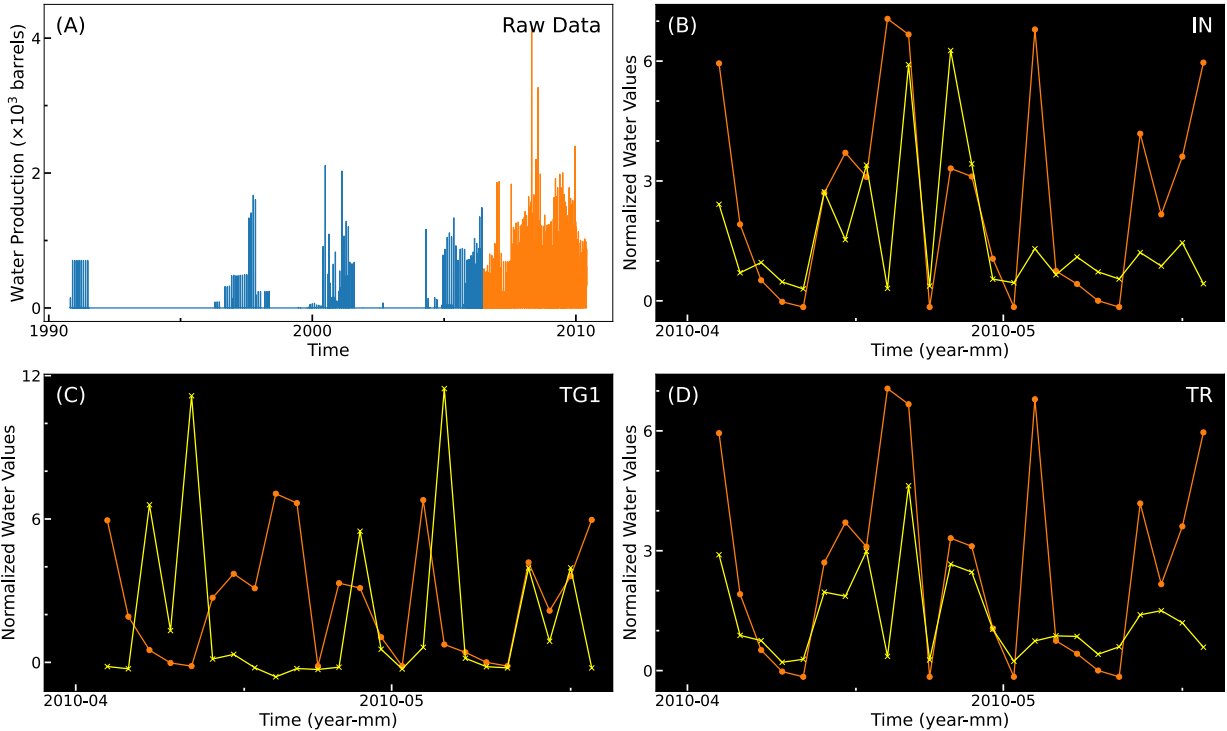

**Figure S4:** Water production forecast for the BEAP site is presented in panels (B, C, D), where yellow lines represent prediction and orange lines denote ground truth. Panel (A) displays the raw water production data, starting from the first non-zero value on October 20, 1990. The blue section indicates training data, and the orange section indicates test data. In panel (B), Informer (IN) results are presented, with 70th percentile chosen as the top performance. Panel (C) displays TimeGrad results at 97th percentile, which underperform due to inadequate training data. Predictions show inconsistency until the beginning of May 2010, after which they align closely with the ground truth. Panel (D) exhibits water data forecast from the Transformer (TR) model at the 50th percentile. Although the Transformer's performance could be enhanced, it demonstrates capability in forecasting water production.

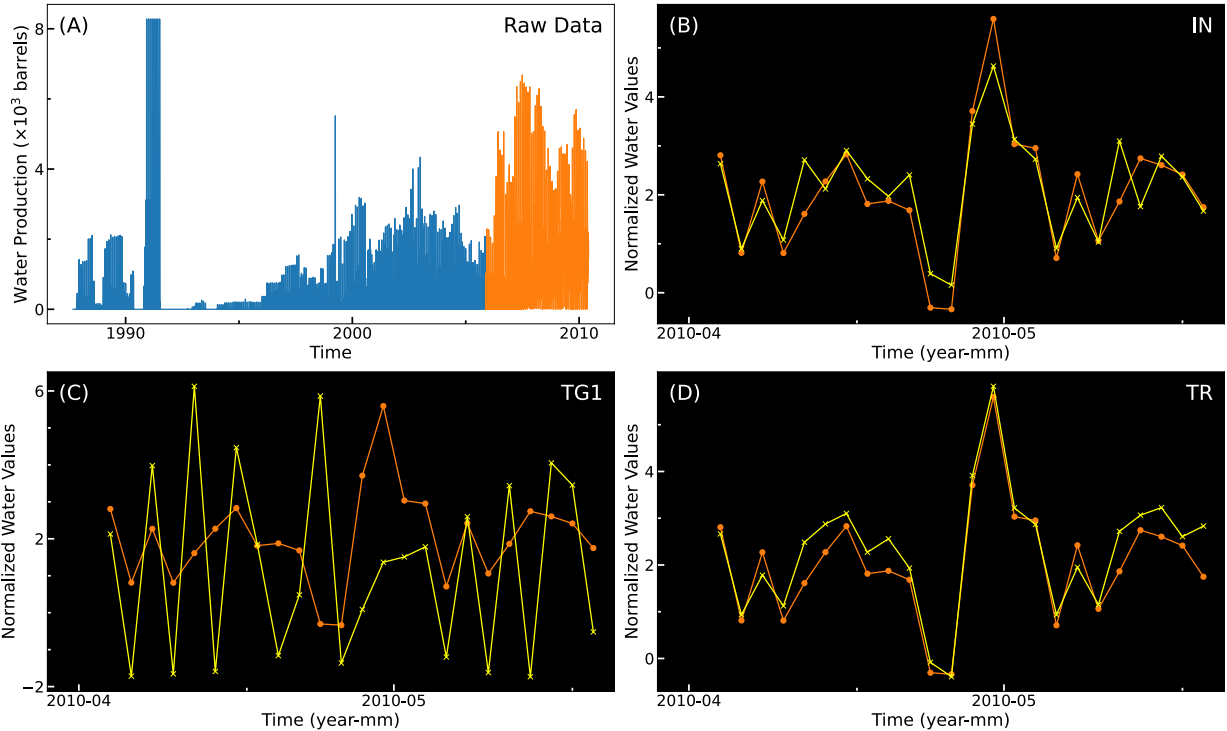

**Figure S5:** Water production forecasts for the BEAT site are presented in panels (B, C, D), where yellow lines represent predictions and orange lines denote ground truth. Panel (A) shows the raw water production data starting from 8 September 1987. The blue section indicates training data, and the orange section indicates test data. In Panel (B), the performance of Informer (IN) at the 50th percentile is depicted. It accurately forecasts the overall data fluctuations. Panel (C) shows that TG1 at the 95th percentile does not perform well in predicting the fluctuations in the raw data. Panel (D) displays the results of the vanilla transformer (TR) at the 50th percentile, which forecasts better than the Informer.

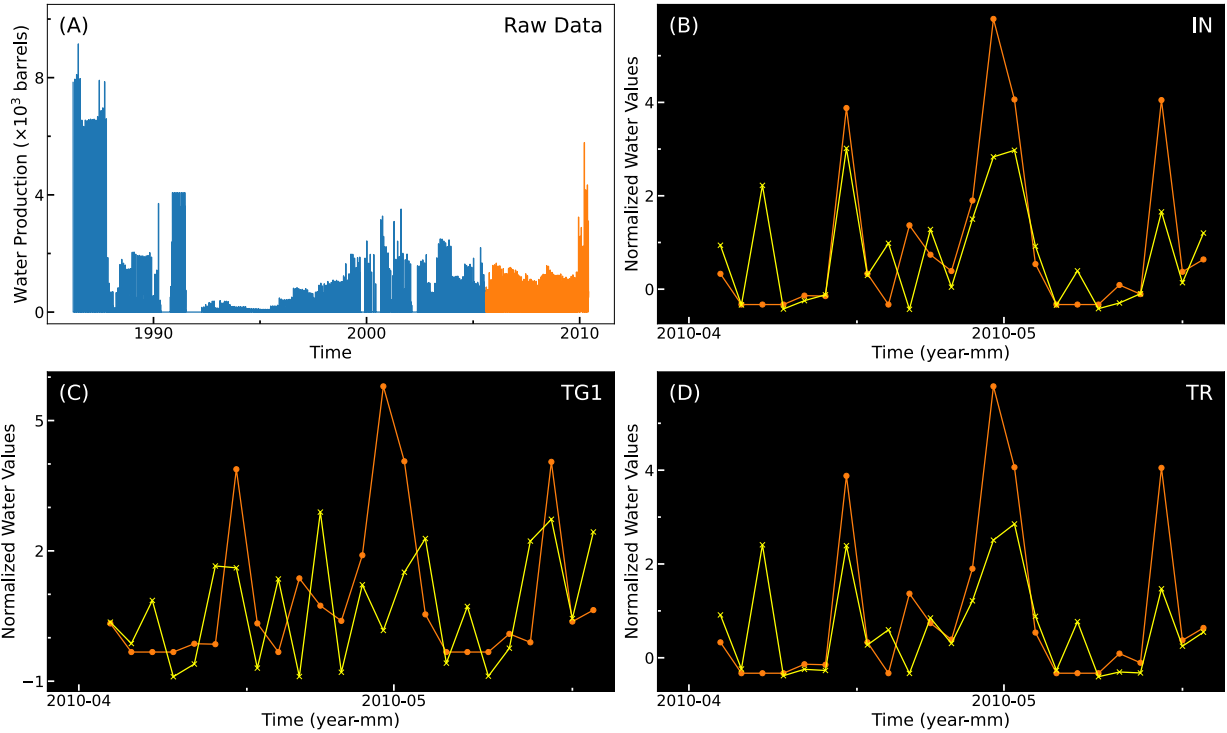

**Figure S6:** Water production forecasts for the EUAT site are presented in panels (B, C, D), where yellow lines represent predictions and orange lines denote ground truth. Panel (A) displays raw water production data starting from 25 March 1986. The raw data is divided into training (blue) and test (yellow) sets. In panel (B), informer (IN) results at the 50$^{th}$ percentile are depicted. IN accurately predicts most of the peak positions. Panel (C) shows the forecast by TG1 at the 90$^{th}$ percentile. Panel (D) illustrates the Transformer (TR) results at the 40$^{th}$ percentile.

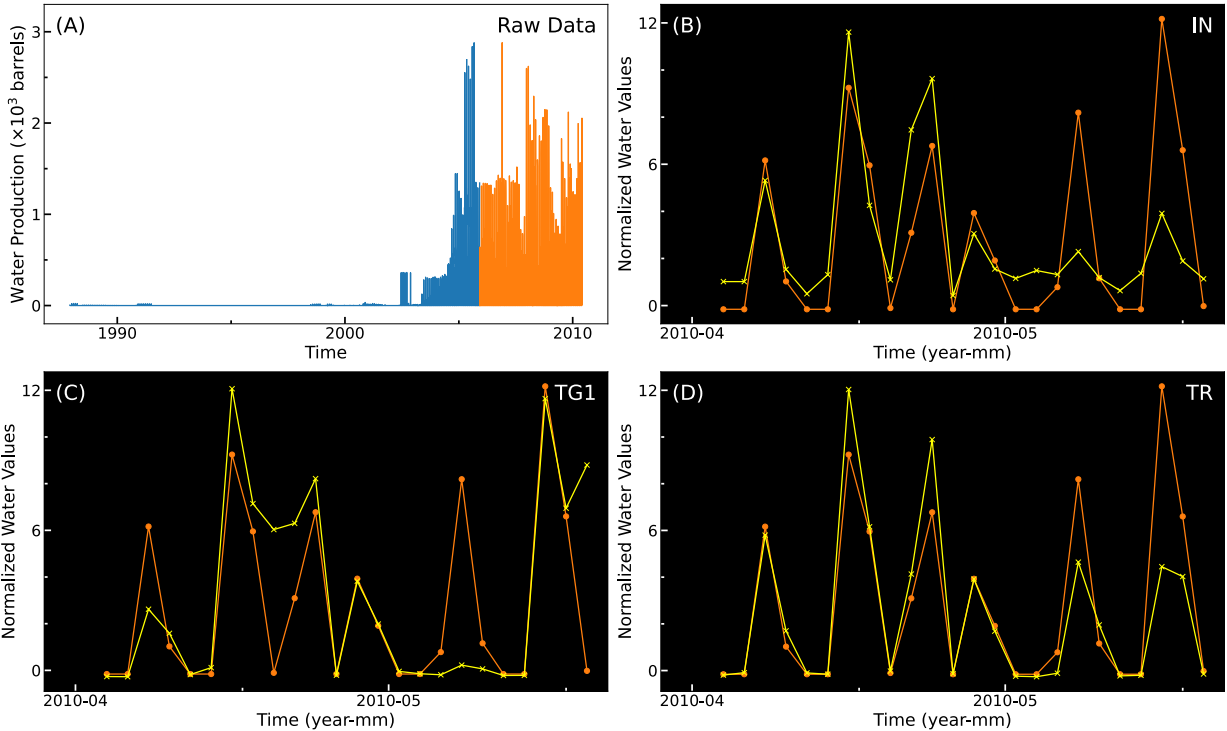

**Figure S7:** Water production forecasts for the EUZE site are presented in panels (B, C, D), where yellow lines represent predictions and orange lines are the ground truth. Panel (A) shows the raw water production data from January 1987 to June 2010. The blue and orange sections represent training and test data, respectively. In panel (B), the performance of the Informer (IN) at the 80$^{th}$ percentile, selected as the best performer, is illustrated. Informer accurately forecasts peak positions but not peak heights. Panel (C) displays TG1 results at the 65$^{th}$ percentile, and panel (D) depicts the forecast of the Transformer (TR) at the 80$^{th}$ percentile. TR accurately predicts most of the peaks and troughs in the ground truth.

**Oil and Water Production Forecast**

Figures S8 and S9 display the oil and water production forecasts of the Informer and TimeGrad models for the BEAT and EUAT sites. For BEAT and EUAT, we merge the oil and water production data. To ensure non-zero values at the outset, we set the training start date for these two sites to the onset of water production at the BEAP site (September 1987), where water breakthrough occurred after EUAT. We use an 80:20 data split for training and testing, forecasting 30 timesteps from April to June 2010. The Informer (IN) is trained for 20 epochs, while the TimeGrad model TG1 undergoes 40 epochs with GRU and L1 loss.

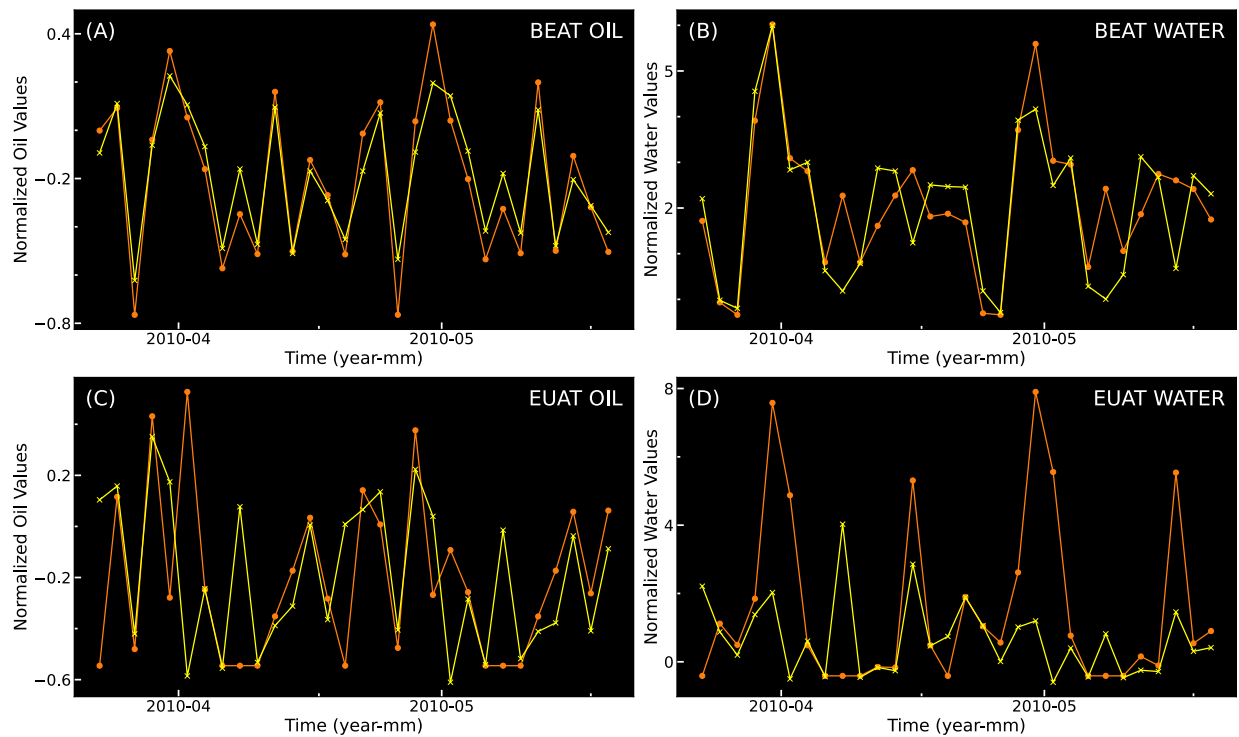

**Figure S8:** Oil and water production forecasts by an Informer model are depicted, merging oil and water data for the BEAT and EUAT sites. In panels (A) and (B), the forecasts of the Informer are illustrated at the 60[th] percentile for oil and 70[th] percentile for water. Panels (C) and (D) illustrate the forecasts for EUAT, respectively, at the 50[th] percentile for both oil and water production. The Informer demonstrates relatively good prediction capability for BEAT compared to the EUAT site.

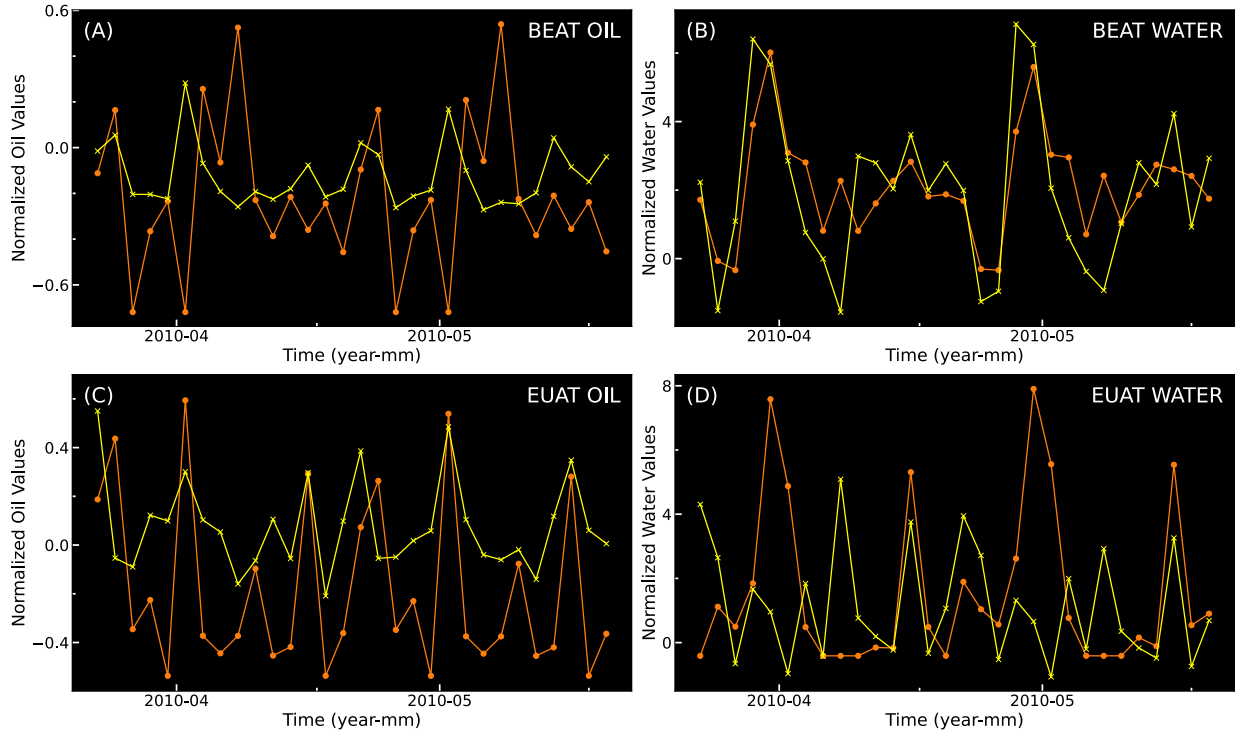

**Figure S9:** TimeGrad production forecasts for oil and water at the BEAT and EUAT sites are presented. Panels (A) and (B) display predictions (yellow) and ground truth (red) for oil and water, respectively, at the BEAT site. The best predictions are at the 95th percentile for oil and the 75[th] percentile for water. Panels (C) and (D) illustrate the same results for EUAT, where we chose 95[th] percentile for oil and the 65[th] percentile for water production.

**Table 1**: Displays Mean Squared Error (MSE) and Mean Absolute Scaled Error (MASE) for oil Production. MSE accounts for both large and small errors, penalizing large errors more heavily. MASE is scale-independent, effectively capturing outliers and extreme values. The Transformer demonstrates overfitting in the forecasts, whereas the Informer provides superior results compared to TimeGrad.

| Models | Sites | MSE | MASE |
|---|---|---|---|
| Only Oil – Multivariate TimeGrad Model | BEAP | 0.041 | 0.359 |
| | BEAT | 0.023 | 0.161 |
| | EUAT | 1.148 | 1.074 |
| | EUZE | 0.433 | 0.825 |
| Oil and Water – Multivariate TimeGrad Model | BEAP | 0.041 | 0.276 |
| | BEAT | 0.025 | 0.187 |
| | EUAT | 0.090 | 0.274 |
| | EUZE | 0.243 | 0.590 |
| Informer | BEAP | 0.011 | 0.193 |
| | BEAT | 0.012 | 0.106 |
| | EUAT | 0.032 | 0.153 |
| | EUZE | 0.044 | 0.262 |
| Vanilla Transformer | BEAP | 0.011 | 0.163 |
| | BEAT | 0.001 | 0.041 |
| | EUAT | 0.038 | 0.136 |
| | EUZE | 0.039 | 0.155 |

**APPENDIX 4: NON-NORMALIZED PREDICTIONS**

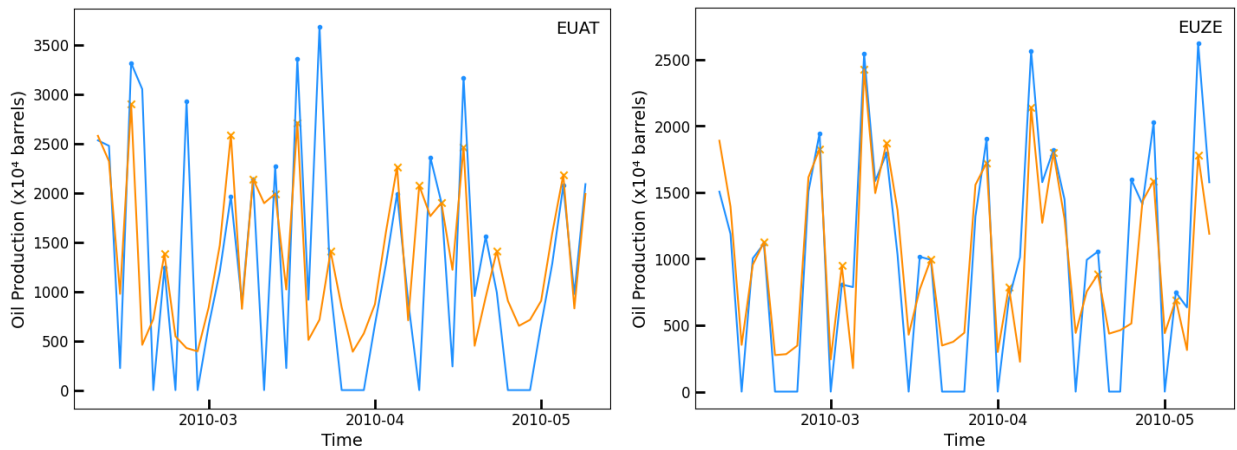

**Figure S10:** Non-normalized forecasts of the Informer model for the sites EUAT and EUZE for 3 months. The predictions are well-aligned with the ground truth.

**APPENDIX 5: OIL PRODUCTION RESULTS FOR 6 MONTHS**

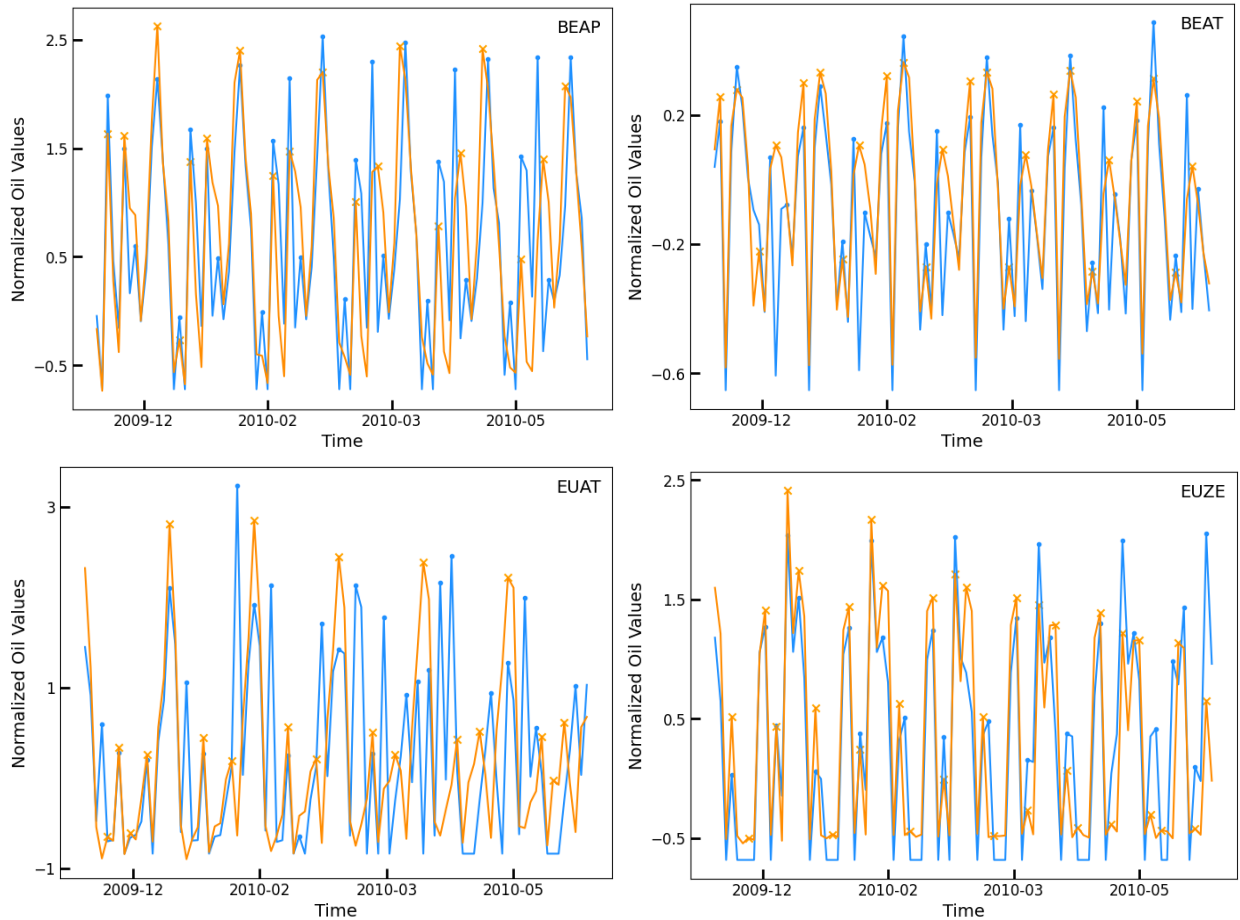

**Figure S11:** The Informer model forecasts for oil production at the sites BEAP, BEAT, EUAT and EUZE. Blue represents the ground truth and the orange represents the predictions generated by the model. The results are for an extended period of 6 months – 90 timesteps. The predicted peaks and troughs align closely with the ground truth for the sites BEAP, BEAT and EUZE. For EUAT, the model can predict the peaks, however, there is some disparity in the magnitude. This is because the oil production declines significantly during the testing phase as compared to the training phase (Figure 3 Panel(A) in the manuscript).

Table 2: Displays the mean and standard deviation for the normalized values of ground truth and the predictions of the last 90 timesteps (6 months) in the test dataset. Across all the sites, the model effectively captures the standard deviation and the mean in the oil production.

| SITE | MEAN GROUND TRUTH | MEAN PREDICTION | STD GROUND TRUTH | STD PREDICTION |
|---|---|---|---|---|
| BEAP | 0.62 | 0.64 | 0.92 | 0.93 |
| BEAT | -0.11 | -0.05 | 0.29 | 0.27 |
| EUAT | 0.24 | 0.18 | 1.01 | 0.97 |
| EUZE | 0.30 | 0.30 | 0.86 | 0.88 |